\documentclass{article}

\usepackage{microtype}
\usepackage{graphicx}
\usepackage{amsmath}
\usepackage{subfigure}
\usepackage{booktabs} 
\usepackage{multirow}
\usepackage{hyperref}

\usepackage[accepted]{icml2021}

\icmltitlerunning{Submission and Formatting Instructions for ICML 2021}

\begin{document}

\twocolumn[
\icmltitle{Urban Tree Species Classification Using Aerial Imagery}



\icmlsetsymbol{equal}{*}

\begin{icmlauthorlist}
\icmlauthor{Emily Waters}{aru}
\icmlauthor{Mahdi Maktabdar Oghaz}{aru}
\icmlauthor{Lakshmi Babu Saheer}{aru}

\end{icmlauthorlist}

\icmlaffiliation{aru}{Faculty of Science and Engineering, Anglia Ruskin University, Cambridge, United Kingdom}

\icmlcorrespondingauthor{Lakshmi Babu Saheer}{lakshmi.babu-saheer@aru.ac.uk}
\icmlcorrespondingauthor{Mahdi Maktabdar}{mahdi.maktabdar@aru.ac.uk}

\icmlkeywords{Machine Learning, ICML}

\vskip 0.3in
]



\printAffiliationsAndNotice{\icmlEqualContribution} 

\begin{abstract}

Urban trees help regulate temperature, reduce energy consumption, improve urban air quality, reduce wind speeds, and mitigating the urban heat island effect. Urban trees also play a key role in climate change mitigation and global warming by capturing and storing atmospheric carbon-dioxide which is the largest contributor to greenhouse gases. Automated tree detection and species classification using aerial imagery can be a powerful tool for sustainable forest and urban tree management. Hence, This study first offers a pipeline for generating labelled dataset of urban trees using Google Map's aerial images and then investigates how state of the art deep Convolutional Neural Network models such as VGG and ResNet handle the classification problem of urban tree aerial images under different parameters. Experimental results show our best model achieves an average accuracy of 60\% over 6 tree species.

\end{abstract}
\section{Introduction}
Trees are well recognised for their importance to the planet and human life. Environmentally, trees slow surface runoff from rainfall, reducing flood risk, water pollution and soil erosion ~\cite{chandler2017influence}. They improve overall air quality by absorbing particulate matter, create a cooling effect, and mitigating the heat island effect in urban areas \cite{manickathan2017parametric}. A study by~\cite{bastin2019global} shows forestation is a possible strategy for mitigating climate change. Trees capture and store atmospheric carbon-dioxide and lock it up for centuries. Trees play a key role in climate change mitigation by capturing, storing and consequently reducing atmospheric CO2 levels, the main adverse contributor to greenhouse gases and climate change. Studies show, urban trees can cut heating costs by reducing wind-speed and casting shade around the housing area  which indirectly mitigates emission of greenhouse gases \cite{wolf2005business}. To leverage this potential, effective forest and urban tree management is essential. This requires detailed information about tree species, composition, health and geographical location of each tree in order to create a long term sustainable plan for plantation and forestation sites, pruning schedules and mitigation of potential problems \cite{baeten2018}. It also helps to monitor tree species diversity and track health and growth rate to creates a more robust ecosystem with better productivity and greater resilience to disease and pests \cite{gamfeldt2013, rust2016tree}. Such management system demands for an accessible, reliable yet economically and practically viable platform to automatically detect, classify and monitor forests and urban trees. Historically, this has been carried out by experts and volunteers visiting trees on the ground but this is a laborious, time-consuming and expensive approach. Alternatively LiDAR technology, used to estimate the number of trees in an area~\cite{wilkes2018estimating} and categorise their species~\cite{kim2008}, paved the way to automated urban tree and forest management. However, LiDAR surveying is a costly process mainly due to the speciality equipment and skilled human resource required to collect and interpret  it~\cite{rezatec20202019}. Hyperspectral imaging and remote sensing satellites images have advanced significantly over the last couple of decades and are now able to produce high-resolution images which facilitates tree detection and species classification \cite{fricker2019convolutional, dalponte2014tree, maschler2018individual, clark2005hyperspectral}. There are a limited number of studies looking into the detection classification of trees using RGB aerial images. RGB aerial image surveying can be as costly as other aforementioned approaches however availability of mapping service such as Google Maps and Bing Maps can significantly reduce the cost of surveying and data collection. Studies like \cite{wegner2020june, nezami2020} utilized images from these platforms paired with Convolutional Neural Network (CNN) to create a fully automated yet accurate tree detection and classification model which is pertinent to effective forest and urban tree management.       

Having said that, the purpose of this study is to first generate a labelled dataset of urban trees using Google Map's RGB aerial images paired with existing tree inventories to supply GPS coordinates and species information. This study uses the Camden tree inventory \cite{coucil_2021} to acquire GPS coordinates and species details. This study also aims to build a supervised model capable to detect and classify tree species accurately. Several state of the art pre-trained CNNs models including VGG and ResNet variants along with some custom models have been investigated, compared and analysed.

\section{Dataset Generation}  
\begin{figure}[h]
\includegraphics[width=8cm]{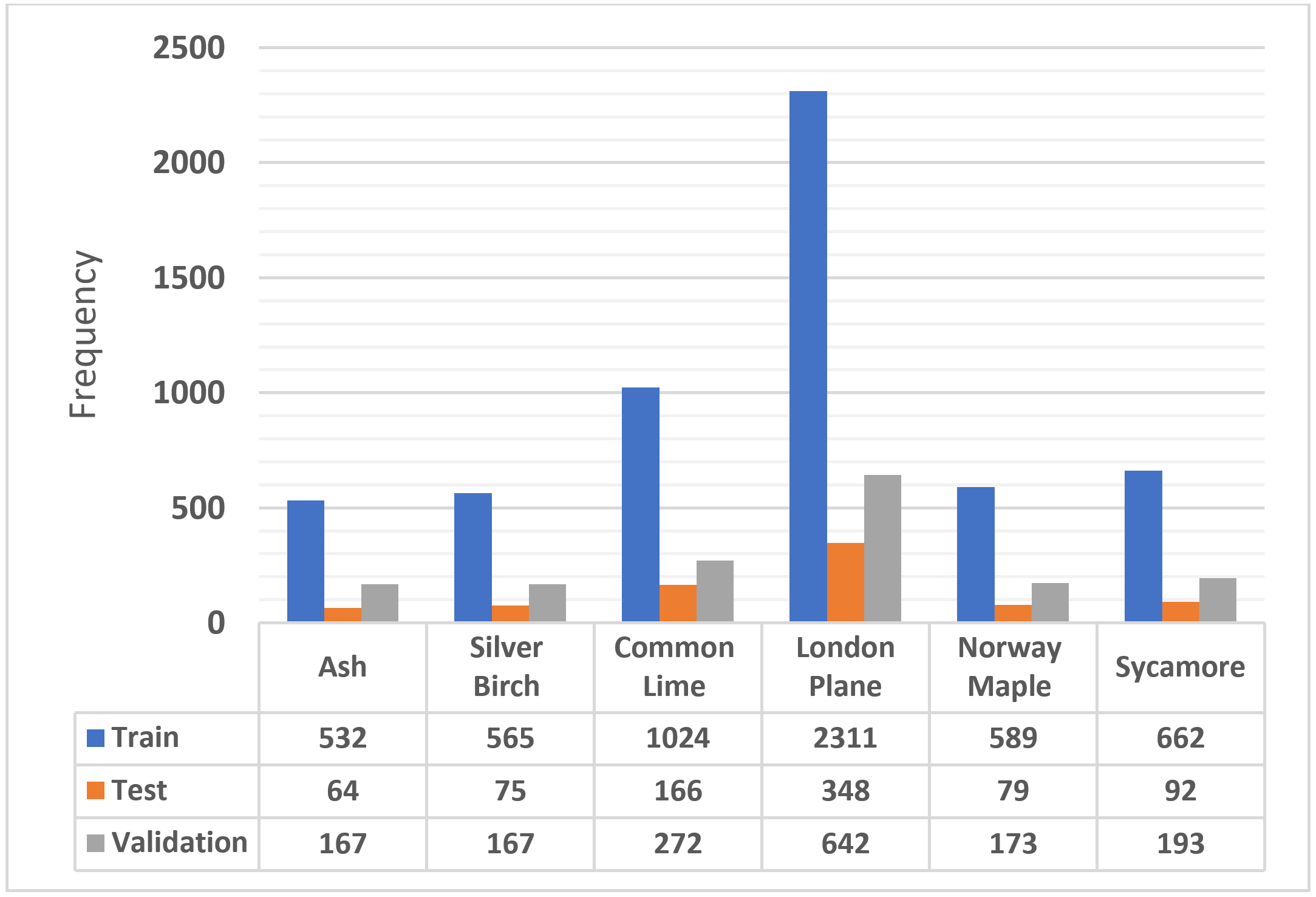}
\caption{Training, validation and testing sets counts across top 6 species in Camden dataset}
\label{fig:speciesfreq}
\end{figure}

\begin{figure}[h]
\includegraphics[width=8cm]{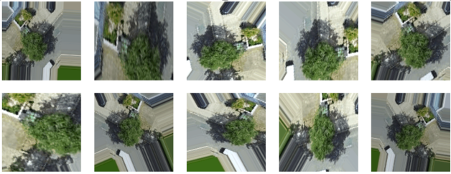}
\caption{Example of the augmentation applied to images in the training data subset}
\label{fig:augmentedpics}
\end{figure}

The proposed dataset generator pipeline uses Google Map's static API to source trees' aerial images and Camden tree inventory \cite{coucil_2021} to supply tree's GPS location and species information. Camden tree inventory contains over 23,000 GPS locations (Latitude and Longitude) of up to date (over 99.9\% of records dated 2016 or later) Council owned trees on highways and in parks and open spaces in London Borough of Camden. Cleaning process performed by removing entries with missing locations, vacant plots or unknown species. Each data point contains tree species, height, spread, diameter at breast height (DBH), and maturity. An automated process, goes through all entries in the Camden inventory and downloads aerial image from Google Map's static API. The latitude and longitude co-ordinates of each tree were used as the centre point for each aerial image of 200x200 and zoom level of 20. While Camden tree inventory consists of hundreds of different tree species, this study only investigates top 6 species with the highest frequencies including Ash, Silver Birch , Common Lime, London Plane, Norway Maple and Sycamore. The data is split into subsets with 70\% for training, 20\% for validation and 10\% reserved for testing. Images were labelled and categorized based on their species and then organized into train, test and validate sub-sets. The proportional representation of each species is preserved across the subsets so that any class imbalance is retained at each stage. As it can be observed in the Figure \ref{fig:speciesfreq}, the number of entries in the training set is fairly limited for an effective train of a deep convolutional model. Hence, this study employed image augmentation technique (Rotation, width and height shift, horizontal flip, zoom and brightness) to over sample and expand the training set with new, plausible examples as shown in the Figure \ref{fig:augmentedpics} \cite{krizhevsky2017imagenet}.


\section{CNN for Tree Species Classification}

This research investigates and evaluates 3 possibilities including VGG-16, ResNet50 and a group of custom deep models to find an optimal CNN model for tree species classification. The VGG-16 ~\cite{simonyan2014very} was the chosen model in similar tree species classification studies by Branson, et al.~\cite{branson2018fromgoogle} and Lang~\cite{lang2020july}. As per these studies, the VGG-16 network was pre-trained with ImageNet dataset~\cite{russakovsky2015e} and then being fine-tuned and optimized on our dataset of tree aerial images. We paired VGG-16 model with Adam optimiser which besides being computationally efficient was also used in similar studies like  Lang~\cite{lang2020july}. The parameters to be varied are dropout and class weights. Class weights applied to compensate for imbalance class sizes. All the models considered in this study are trained and tested based on the training, validation and testing sets shown in the Figure \ref{fig:speciesfreq}.  This study used categorical cross-entropy loss function across all models in this study while optimiser choice varied to see which has the greatest impact on model performance. The maximum number of training epochs is set to 100. During training, the model with the smallest loss is saved and used for comparison with other models. To reinforce the evaluation process, the top 5 models with the smallest loss and higher accuracy have further evaluated using 5-fold cross validation to obtain more reliable results. 
This study also investigates the performance of pre-trained ResNet50 model for tree species classification using aerial images. Many similar studies including~\cite{natesan2019resnet, cao2020improved} used this model for similar purposes. Deep structure of Resnet50 facilitates modeling of complex features while skip connections avoid issues like vanishing gradients.ResNet50 has been paired with Adam optimizer to achieve efficient training and timely convergence. Various dropout and class weight ratios have been examined and optimized during the training process. In addition to aforementioned pre-trained models, this study investigates a range of Custom CNNs models to identify possibility of achieving accurate tree species classification using a less complex model. The template for construction of these custom models is illustrated in equation~\ref{equation:constructedcnn}.

\begin{equation}
\begin{split}
& INPUT \rightarrow \\
& [[CONV \rightarrow RELU]*2 \rightarrow MAXPOOL]*N \rightarrow \\
& [FC \rightarrow RELU] \rightarrow FC, \\
& where N\in\{1, 2, 3, 4, 5, 6\}
\end{split}
\label{equation:constructedcnn}
\end{equation}

where N is the number of Convolutional blocks, ranges between 1 and 6, with each block consisting of two Convolutional layers (CONV) with a ReLU activation function  followed by a Maxpooling layer. The size of the kernel and choice of kernel initialiser within the CONV layers are to be varied between models. Dropout is added after each Convolutional block and after the penultimate fully connected (FC) layer. Optimiser choice varied to identify its impact on model performance.

\section{Results and Discussion} 

The training process is conducted using the 6 tree species (Ash, Silver Birch , Common Lime, London Plane, Norway Maple  and Sycamore) with the largest number of samples. The VGG-16, ResNet50 and a range of Custom CNNs have been trained with a combination of different parameters including dropout ratio, optimiser, class balanced weight to identify the top performer model. A further model re-evaluation using 5-fold stratified cross validation helped to obtain more reliable accuracy figures.

The VGG-16 model is trained with various dropout and class weights values. The performance measures obtained by the VGG-16 model are presented in Table~\ref{table:vgg16results}. The VGG-16 base model achieves an accuracy of 56.55\%, only outperformed by the VGG-16 model with 20\% dropout that increases the score by 0.61\%. The accuracy differences for the VGG-16 models with or without dropout appear to be marginal, however the precision gains almost 5\% for the model with 20\% dropout. Note that the class balanced weight model under-performs other models with a considerable margin.  Similar to VGG-16, ResNet50 model is trained with various dropout and class weights values. The performance measures obtained by the ResNet50 model are recorded in Table~\ref{table:resnet50results}. The standard ResNet50 model managed to achieve accuracy of 59.03\% which is already higher than any figure achieved by the VGG-16 model. Adding a 20\% dropout, marginally raised ResNet50 accuracy to 59.92\%. Moreover, Average Class Precision raised by almost 3\% for the model with 20\% dropout. Other models including Balanced class weights and 10\% dropout perform more or less the same as the standard ResNet50 model. 

\begin{table}
\centering
\caption{Comparison of results for the VGG-16 variants}
\resizebox{\linewidth}{!}{%
\begin{tabular}{llllll} 
\toprule
Model & Loss & Accu (\%) & \begin{tabular}[c]{@{}l@{}}Ave Class\\Recall (\%)\end{tabular} & \begin{tabular}[c]{@{}l@{}}Ave Class\\Precision (\%)\end{tabular} & No Epochs \\ 
\hline
\begin{tabular}[c]{@{}l@{}}VGG-16\\(Standard)\end{tabular} & 1.1934 & 56.55 & 42.23 & 40.94 & 68 \\
\begin{tabular}[c]{@{}l@{}}VGG-16\\(Balanced W)\end{tabular} & 1.7900 & 42.23 & 16.67 & 7.04 & 13 \\
\begin{tabular}[c]{@{}l@{}}VGG-16\\(10\% dropout)\end{tabular} & 1.1914 & 55.83 & 41.80 & 40.30 & 49 \\
\begin{tabular}[c]{@{}l@{}}VGG-16\\(20\% dropout)\end{tabular} & 1.1649 & 57.16 & 42.65 & 45.30 & 88 \\
\bottomrule
\end{tabular}
}
\label{table:vgg16results}
\end{table}

\begin{table}
\centering
\caption{Comparison of results for the ResNet50 variants}
\resizebox{\linewidth}{!}{%
\begin{tabular}{llllll} 
\toprule
Model & Loss & Accu (\%) & \begin{tabular}[c]{@{}l@{}}Ave Class\\Recall (\%)\end{tabular} & \begin{tabular}[c]{@{}l@{}}Ave Class\\Precision (\%)\end{tabular} & No Epochs \\ 
\hline
\begin{tabular}[c]{@{}l@{}}ResNet50\\(Standard)\end{tabular} & 0.86 & 59.03 & 51.13 & 49.37 & 43 \\
\begin{tabular}[c]{@{}l@{}}ResNet50\\(Balanced W)\end{tabular} & 1.03 & 58.96 & 50.66 & 48.82 & 41 \\
\begin{tabular}[c]{@{}l@{}}ResNet50\\(10\% dropout)\end{tabular} & 0.88 & 59.14 & 51.24 & 49.44 & 46 \\
\begin{tabular}[c]{@{}l@{}}ResNet50\\(20\% dropout)\end{tabular} & 0.73 & 59.92 & 54.07 & 52.46 & 62 \\
\bottomrule
\end{tabular}
}
\label{table:resnet50results}
\end{table}

Apart from pre-trained VGG-16 and ResNet50, we have trained and evaluated a range of Custom CNNs models to identify possibility of achieving accurate tree species classification using a less complex model. All custom models are constructed as per the formula in equation~\ref{equation:constructedcnn}. 
The baseline custom model has one convolutional block with 3x3 kernels, "He uniform" initialiser and SGD optimiser. This model is then compared with a few other custom models which mainly differ by having 2 to 6 convolutional blocks, different dropout ratio, kernel size, optimizer and initialiser. A detailed summary of notable results are presented in the Table~\ref{table:customCNN}. The choice of optimiser was limited to what Tensorflow library offers. We have explored different optimisers including Adadelta, Adagrad, Adam, Adamax, Ftrl, Nadam, RMSprop and SGD. Experiment results shows the Adamax optimiser consistently outperformed other optimisers in this comparison. Similarly, the initialisers are taken from the Tensorflow offerings including constant, Glorot normal, Glorot uniform, He normal, He uniform, Lecun normal, Lecun uniform and random normal. Results shows "He normal" marginally outperforms other initialisers in this comparison. According to the Table~\ref{table:customCNN}, The top performing model has 6 convolutional blocks paired with the “He normal” kernel initialiser and is optimised using Adamax – a variant of the Adam algorithm. This model achieves accuracy of 69.5\%, recall of 57.4\% and precision of 62.8\%. The top performing model re-evaluate using 5-fold stratified cross validation which led to a considerable drop across majority of the metrics. Cross validation Result can be observed at the bottom row of the Table \ref{table:customCNN}. Qualitative results of the top model can be found in the Appendix 1.

\begin{table}
\centering
\caption{Comparison of results for custom CNN models}
\resizebox{\linewidth}{!}{%
\begin{tabular}{llllll} 
\toprule
Model & Loss & Accu (\%) & \begin{tabular}[c]{@{}l@{}}Ave Class\\Recall (\%)\end{tabular} & \begin{tabular}[c]{@{}l@{}}Ave Class\\Precision (\%)\end{tabular} & No Epochs \\ 
\hline
\begin{tabular}[c]{@{}l@{}}x1 Conv block\\(Baseline)\end{tabular} & 1.2495 & 52.79 & 35.99 & 38.37 & 44 \\
\begin{tabular}[c]{@{}l@{}}x2 Conv block\end{tabular} & 1.1929 & 55.95 & 38.53 & 40.55 & 62 \\
\begin{tabular}[c]{@{}l@{}}x3 Conv block\end{tabular} & 1.1219 & 58.50 & 42.81 & 54.06 & 57 \\
\begin{tabular}[c]{@{}l@{}}x3 Conv block\\(20\% dropout)\end{tabular} & 1.2326 & 54.98 & 40.12 & 42.27 & 48 \\
\begin{tabular}[c]{@{}l@{}}x3 Conv block\\(30\% dropout)\end{tabular} & 1.2604 & 51.46 & 33.49 & 36.33 & 67 \\
\begin{tabular}[c]{@{}l@{}}x4 Conv block\end{tabular} & 1.1644 & 58.13 & 42.81 & 45.54 & 35 \\
\begin{tabular}[c]{@{}l@{}}x5 Conv block\end{tabular} & 1.0752 & 62.37 & 49.37 & 52.89 & 54 \\
\begin{tabular}[c]{@{}l@{}}x3 Conv block\\(5x5 Kernel)\end{tabular} & 1.2031 & 54.98 & 39.17 & 44.19 & 28 \\
\begin{tabular}[c]{@{}l@{}}x3 Conv block\\(Adam)\end{tabular} & 1.0072 & 65.53 & 49.38 & 52.89 & 49 \\
\begin{tabular}[c]{@{}l@{}}x3 Conv block\\(Glorot uniform)\end{tabular} & 1.2121 & 55.58 & 40.73 & 45.61 & 97 \\
\begin{tabular}[c]{@{}l@{}}x6 Conv block\\(adamax he\_normal)\end{tabular} & 0.8836 & 69.54 & 57.41 & 62.75 & 58 \\
\begin{tabular}[c]{@{}l@{}}x6 Conv block\\(adamax lecun\_normal)\end{tabular} & 0.9299 & 69.17 & 58.00 & 62.54 & 69 \\
\begin{tabular}[c]{@{}l@{}}x5 Conv block\\(adamax glorot\_normal)\end{tabular} & 0.9088 & 69.17 & 57.16 & 61.67 & 75 \\
\begin{tabular}[c]{@{}l@{}}x5 Conv block\\(adamax truncated\_normal)\end{tabular} & 0.9418 & 68.33 & 55.43 & 61.75 & 69 \\
\begin{tabular}[c]{@{}l@{}}x6 Conv block\\(adamax he\_uniform)\end{tabular} & 0.9228 & 67.72 & 55.10 & 56.82 & 46 \\
\begin{tabular}[c]{@{}l@{}}x5 Conv block\\(adamax he\_normal)\end{tabular} & 0.9254 & 67.11 & 55.74 & 58.97 & 55 \\
\begin{tabular}[c]{@{}l@{}}x6 Conv block\\(adamax truncated\_normal)\end{tabular} & 0.9490 & 66.75 & 53.31 & 58.93 & 85 \\
\begin{tabular}[c]{@{}l@{}}x6 Conv block\\(adamax lecun\_uniform)\end{tabular} & 0.9440 & 66.14 & 51.56 & 60.76 & 61 \\
\begin{tabular}[c]{@{}l@{}}x5 Conv block\\(nadam he\_normal)\end{tabular} & 0.9882 & 66.14 & 51.39 & 59.57 & 59 \\
\begin{tabular}[c]{@{}l@{}}x6 Conv block\\(adagrad he\_uniform)\end{tabular} & 0.9633 & 65.53 & 50.37 & 58.22 & 70 \\
\hline
\begin{tabular}[c]{@{}l@{}}x6 Conv block\\(adamax he\_normal)\\5-fold Cross Val\end{tabular} & -NA- & 60.29 & 46.57 & 56.18  & 100 \\
\bottomrule
\end{tabular}
}
\label{table:customCNN}
\end{table}

\section{Discussion}
\label{sec:discussion}

The VGG-16 network (with up to 20\% dropout) can identify tree species accurately 56\% of the time. The literature indicated that the VGG-16 architecture would generalise well to new classification problems and benefit from being pre-loaded with ImageNet weights. However, our custom CNN models with 3 or more convolutional blocks, consistently outperformed the VGG-16 variants that had been trained. The ResNet50 performed slightly better than the VGG-16 however its performance was inferior to our custom made models. This was a surprising result, perhaps indicating that VGG-16 and ResNet50 were over complex for the task which negatively impacted generalization. The top performing custom CNN model consists of 6 convolutional blocks paired with the Adamax optimiser and He Normal kernel initialiser, which achieved 69\% accuracy on the test set and average 60.29 accuracy on 5-fold stratified cross validation. The Adamax optimiser was a common parameter across the most successful 8 models. Datasets with many outliers or which are noisy in terms of gradient updates can benefit from Adamax over Adam~\cite{kingma2014adam}. Adamax is a sparse implementation of Adam~\cite{kingma2014adam} and in this case was shown to be superior. The VGG-16 and ResNet50 models were only optimised using Adam, so future experiments could explore the effects of using different optimisers here too. Other known architectures such as AlexNet could be trialled in addition to VGG-16 and ResNet50. More convolutional layers increase the number of parameters and have the effect of allowing the model to extract more features – up to a point – after which overfitting tends to occur. This could be a reason for the VGG-16 or ResNet50 networks failing to achieve superior results. Amongst the top models, a pattern emerged to identify that kernel initialisers with normal distributions tended to outperform uniform distributions but its impact was almost negligible.  The strategy for constructing a custom CNN could be extended to explore other possibilities such as altering convolutional blocks to contain 3 convolutional layers instead of 2, varying other parameters such as batch size and learning rate. We realized 5-fold stratified cross validation led to a considerable drop across majority of the metrics. This implies that our dataset is not large and homogeneous enough to generate reliable results in hold-out test method. Another reason for the disparity could be that the 5-fold validation process uses 20\% of the data for each model to be tested on, whereas 10\% is retained for the hold-out test method.

Further investigation shows the top performer model struggles at identifying some tree species such as Ash. We believe this is mainly due to limited number of training samples. This could be mitigated by setting up a hierarchical tree species classification model where a top-level model classifies tree's species family while a separate sub-model will be trained to distinguish between each species of a family. Alternatively, ensemble modelling could be employed – where several models are trained on the data, and their predictions are aggregated to produce a final prediction.

\section{Conclusion}
\label{sec:conclusion}

This work examined the possibility of generating a labelled tree species dataset using Google Maps aerial images and publicly available tree inventories to supply GPS coordinates and tree species information. Moreover, this study offered a deep convolutional neural network model capable to successfully classify tree species using the proposed dataset. The work involved looking at both transfer learning approach using the VGG-16 and ResNet networks and constructing a series of custom CNN models. The top performer model in this research managed to classify up to 6 different tree species with over 60\% average accuracy. Future work such as investigating other pre-trained models under different parameters could likely to improve the metrics. Furthermore, genetic algorithm technique could be adopted to optimise and evolve model parameters and identify the best performing architecture.

\bibliography{example_paper}
\bibliographystyle{icml2021}

\onecolumn
\clearpage
\section{Appendices}
\textbf{Appendix 1}

\begin{figure*}[h]
  \includegraphics[width=\textwidth]{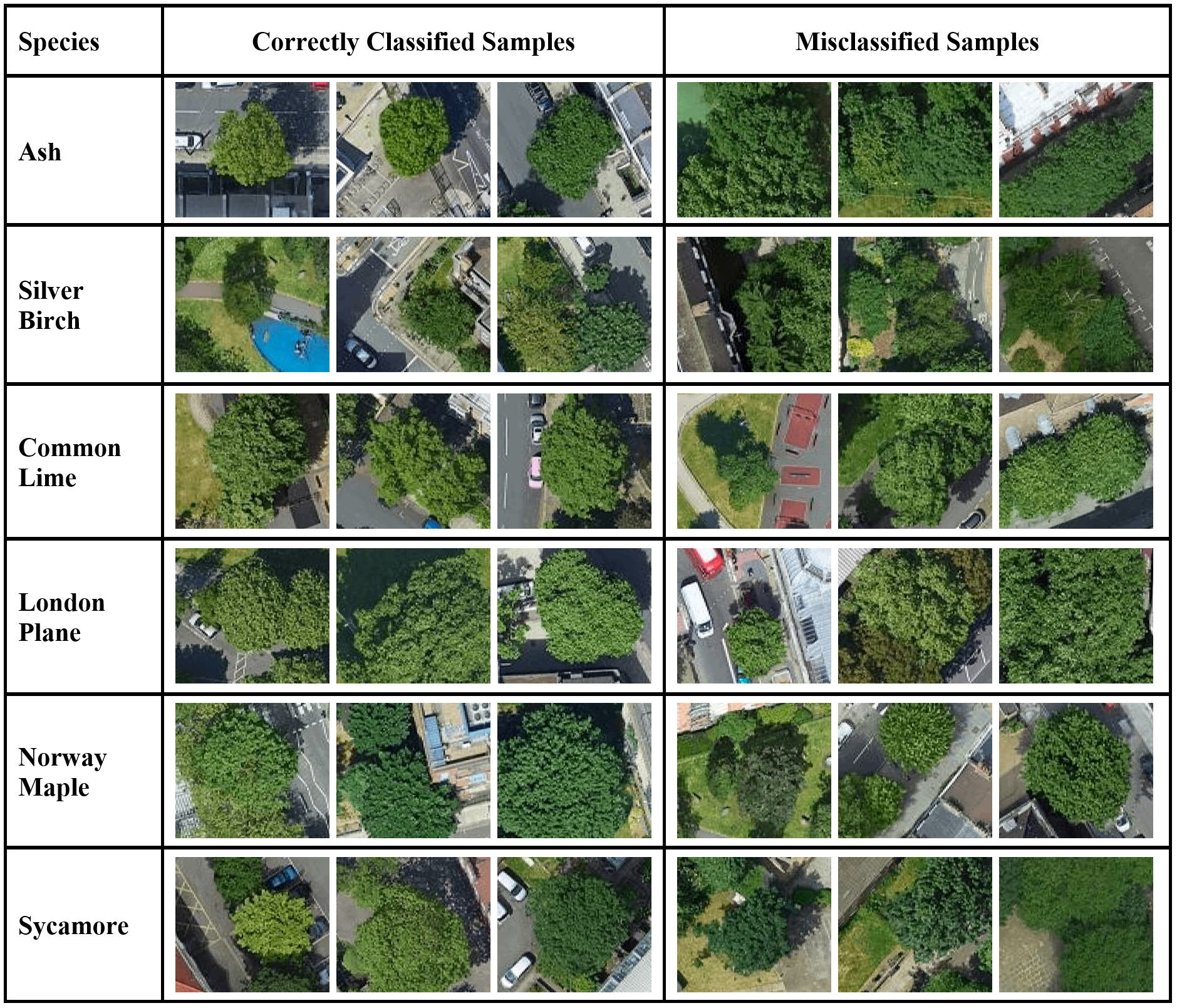}
 Qualitative results generated by the top performer model with the 6 convolutional block, Adamax optimiser and He normal kernel initialiser
\end{figure*}

\end{document}